%% file: acl_latex.tex
\pdfoutput=1

\documentclass[11pt]{article}

\usepackage[preprint]{acl}

\usepackage{times}
\usepackage{latexsym}
\usepackage{booktabs}
\usepackage{subcaption}
\usepackage{pgfplots}
\pgfplotsset{compat=1.18}
\usepackage{xcolor}
\usepackage{tikz}
\usetikzlibrary{arrows.meta}
\usepackage{url}

\usepackage[T1]{fontenc}

\usepackage[utf8]{inputenc}

\usepackage{microtype}

\usepackage{inconsolata}

\usepackage{graphicx}
\usepackage{amsmath}
\usepackage[capitalize, noabbrev]{cleveref}

\usepackage{enumitem}

\newtheorem{definition}{Definition}

\definecolor{dark1}{RGB}{27,158,119}
\definecolor{dark2}{RGB}{217,95,2}
\definecolor{dark3}{RGB}{117,112,179}
\definecolor{dark4}{RGB}{231,41,138}

\tikzset{
	probability bar/.style = {fill=#1, draw=#1!50!black, thin},
	verb label/.style = {font=\footnotesize, anchor=north, draw=none, inner sep=1pt},
	tick label/.style = {font=\footnotesize, anchor=east, inner sep=1pt},
}

\newlength{\probabilityheight}
\NewDocumentCommand\labeledprobbar{m m m m}{
	\begin{scope}[shift={#4}]
		\fill[probability bar=#2] (-4.5pt, 0) rectangle +(9pt, #3);
		\node[verb label] {#1\strut};
	\end{scope}
}

\NewDocumentCommand\modernaxis{m m}{
	\begin{scope}[shift={#2}]
		\draw[thick] (0, 0) -- (0, \probabilityheight);
		\foreach \p/\n in {#1}{
			\draw[thick] (0, \p*\probabilityheight) -- +(-3pt, 0);
			\node[tick label] at (-3pt, \p*\probabilityheight) {\n};
		}
	\end{scope}
}

\NewDocumentCommand\verbdistribution{m m m}{
	\begin{scope}[shift={#3}]
		\foreach \x [count=\i] in {#2}{
			\labeledprobbar{\(v_\i\)}{dark\i}{\x*\probabilityheight}{(10pt*\i, 0)}
		}
		\modernaxis{0/0\%, 0.25/25\%, 0.5/50\%, 0.75/75\%, 1/100\%}{(0, 0)}
		\node[anchor=south, rotate=90, inner sep=0pt] at (-25pt, 0.5\probabilityheight) {#1};
		\node[anchor=north, inner sep=0pt] at (25pt, -12pt) {verbs};
	\end{scope}
}

\pgfplotsset{
        modern/.style={
                enlargelimits=false,
                separate axis lines,
                semithick,
                axis x line*=bottom,
                axis x line shift=10pt,
                axis y line*=left,
                axis y line shift=10pt,
                every axis/.append style={thick},
                tick style={thick, black},
                tick align=outside,
        }
}

\title{Systematic Generalization in Language Models Scales with Information Entropy}

\author{Sondre Wold, Lucas Georges Gabriel Charpentier, Étienne Simon \\
Language Technology Group \\
University of Oslo}

\begin{document}
\maketitle
\begin{abstract}
Systematic generalization remains challenging for current language models, which are known to be both sensitive to semantically similar permutations of the input and to struggle with known concepts presented in novel contexts. Although benchmarks exist for assessing compositional behavior, it is unclear how to measure the difficulty of a systematic generalization problem. In this work, we show how one aspect of systematic generalization can be described by the entropy of the distribution of component parts in the training data. We formalize a framework for measuring entropy in a sequence-to-sequence task and find that the performance of popular model architectures scales with the entropy. Our work connects systematic generalization to information efficiency, and our results indicate that success at high entropy can be achieved even without built-in priors, and that success at low entropy can serve as a target for assessing progress towards robust systematic generalization. 
\end{abstract}

\section{Introduction}%
\label{section:introduction}
\input{sections/introduction}%

\section{Background}%
\label{section:background}
\input{sections/background}%

\section{Measuring Systematicity}%
\label{section:measuring}
\input{sections/measuring_systematicity}
\section{Experiments}%
\label{section:experiments}
\input{sections/experiments}%

\section{Previous Work}%
\label{section:previous_work}
\input{sections/previous_work}%

\section{Conclusion}%
\label{section:conclusion}
\input{sections/conclusion.tex}%

\section*{Limitations}%
\input{sections/limitations}
\section*{Acknowledgements}%
\input{sections/acknowledgements}%

\bibliography{anthology,custom}

\newpage
\appendix

\section{Modified SCAN grammar}
\label{appendix:scan_grammar}
\begin{table}[htb]
    \centering
    \begin{tabular}{ll}
        \toprule
        $V$ & $A$\\
        \midrule
        look & left \\
        jump & right \\
        run & opposite left \\
        walk & opposite right \\
        sprint & around left\\
        crawl & around right\\
        squat & twice \\
        lunge & thrice\\
        \bottomrule
    \end{tabular}
    \caption{The verbs and adjective phrases in our modified version of SCAN.}
    \label{appendix:tab:verbs}
\end{table}

Our modified SCAN grammar consists of the same parts of speech as the original, but we extend the set of verbs $V$ with four additional items. We keep the original eight adjective phrases. The verbs and adjective phrases can be seen in \cref{appendix:tab:verbs}. Every verb can be combined with every adjective phrase in each embedded sequence (under no distributional constraints), combined with one of the two conjunctive phrases from SCAN (and, after).  

\section{Model details}
\label{appendix:model_details}

All models were implemented in Pytorch \citep{paszke2019pytorch}. We conducted hyperparameter searches for each model per experiment. Below is a list of the search space used for each. All configurations used a weight decay of $1 \times 10^{-1}$ and a 10\% dropout rate.

\subsection{RNN}
\label{appendix:model_details:rnn}
We use the basic sequence-to-sequence implementation from \citet{gordon2019permutation}, which is based on the Pytorch MT tutorial. We conduct a grid search over the following learning rates: $\{1 \times 10^{-3}, 1 \times 10^{-4}, 3 \times 10^{-4}\}$; cell type: Elman RNN \citep{elman1991distributed}, LSTM \citep{lstm}, GRU \citep{chung2014empirical}; number of hidden units: $\{64, 128, 256\}$; number of layers: $\{1, 2, 3\}$; and teacher-forcing rate: $\{0.0, 0.5 \}$. All models ran with an attention mechanism between the encoder and decoder, bidirectional layers, and with no learning rate decay. Following previous work, we used a batch size of 1.

The RNN used for the experiment outlined in \cref{sec:experiment_1} has a learning rate of $1 \times 10^{-4}$, normal RNN cell types, a hidden size of 128 and 2 layers (483\,438  parameters), while for the experiment outlined in \cref{sec:experiment_2} it has 1 layer and 64 hidden units (112\,238 parameters). Teacher-forcing did not result in increased performance for either experiment. 

\subsection{CNN}
\label{appendix:model_details:cnn}

We based our implementation of the CNN from \citet{gehring2017convolutional} on the one found in \textsc{fairseq} \citep{ott2019fairseq}. We conduct a grid search over the following learning rates: $\{1 \times 10^{-2}, 1 \times 10^{-3}, 1 \times 10^{-4}, 3 \times 10^{-4}\}$; kernel size: $\{3, 5\}$; number of hidden units: $\{64, 128, 256\}$; and number of layers: $\{1, 2, 3\}$; We train with a batch size of 32 and a cosine decay on the learning rate. 

The CNN used for the experiment outlined in \cref{sec:experiment_1} has a learning rate of $3 \times 10^{-4}$, a kernel size of 5, a hidden size of 64 and 3 layers (298\,078 parameters), and the same configuration for \cref{sec:experiment_2}.

\subsection{Transformer}
\label{appendix:model_details:transformer}
We implement the Transformer following the original from \citet{vaswani2017attention}. We conduct a grid search over the following learning rates: $\{1 \times 10^{-3}, 1 \times 10^{-4}, 3 \times 10^{-4}, 1 \times 10^{-5}\}$; number of hidden units: $\{64, 128, 256\}$; and number of layers: $\{1, 2, 3\}$; As for the CNN, we train with a batch size of 32 and a cosine decay on the learning rate. 

The Transformer used for the experiment outlined in \cref{sec:experiment_1} has a learning rate of $3 \times 10^{-4}$, a hidden size of 128 and 3 layers (1\,796\,894 parameters), while for the experiment outlined in \cref{sec:experiment_2} it has 2 layers and 256 hidden units (4\,763\,166 parameters).

\subsection{Positional Embedding}
\label{appendix:model_details:positional}
In addition to the absolute encodings used in \citet{vaswani2017attention}, we implement two other variations of positional embeddings: RoPE \citet{su2024roformer} and Disentangled \citet{he2021deberta}. For all variations, we use the same hyperparameters found in the Transformer search (\cref{appendix:model_details:transformer}). Additionally, we conduct a hyperparameter search for the encoding-specific parameters. For RoPE we conduct a search for the base value of the $\theta$. We tried powers of 10, from one to one million. For Disentangled Attention we sweep over the input length for the relative attention. We have a separate length for the input and output encodings. We sweep over the following values: $\{4, 8, 16, 32, 64\}$. As for the Transformer, we train with a batch size of 32 and a cosine decay on the learning rate.

For the models used for the experiment outlined in \cref{sec:experiment_position_embedding}, we use a RoPE $\theta$ value of 1\,000, and for the disentangled encodings we use a relative input length of 64 and a relative output length of 4.

\subsection{Permutation equivariant sequence model}
\label{appendix:model_details:equivariant_model}
For the permutation equivariant model, we use the original implementation from \citet{gordon2019permutation}. We extend the cyclic group from four to eight verbs. For the hyperparameter selction, we conducted a grid search over the following learning rates: $\{1 \times 10^{-3}, 1 \times 10^{-4}, 3 \times 10^{-4}, 1 \times 10^{-5}\}$; the number of hidden units: $\{64, 128\}$; and RNN cell type: Elman RNN, GRU. All settings ran with 1 layer. 

The model used for the experiment outlined in \cref{sec:experiment_1} has a learning rate of $1 \times 10^{-4}$, a hidden size of 64 and the GRU cell types ($\approx$ 130\,000 parameters), and the same configuration for \cref{sec:experiment_2}.

We note that the model presented in \citet{lake2023human} also encodes verb equivariance and achieves high performance on SCAN. However, this approach relies on automatically augmenting the training data by creating examples of the input that have different semantics than the original grammar, making our quantification of $H$ impractical. Consequently, we only evaluate the permutation-equivariant model from \citet{gordon2019permutation}.




\section{Computational budget}
\label{appendix:computational_budget}

All hyperparameter sweeps and experiments ran on AMD EPYC 7763 CPUs:

\begin{itemize}[nosep]
    \item RNN sweep:  2880 hours.
    \item CNN sweep: 139 hours.
    \item Transformer sweep: 120 hours.
    \item Permutation equivariant sweep: 24 hours.
    \item RNN training run: 600 hours.
    \item CNN training run: 21 hours.
    \item Transformer sweep: 120 hours.
    \item Evaluation: 12 hours.
    \item \textbf{Total:} 3916 core CPU hours.
\end{itemize}

\section{Licenses and use}
\label{appendix:licenes}
The present work uses the following scientific artifacts: 

\begin{enumerate}
    \item We base our synthetic data on the SCAN grammar \citep{lake2018generalization}. SCAN is released under a BSD License. We do not redistribute or use any of the source code, nor do we redistribute or use any of the original data files. We implement the grammar based on the description provided in the paper. Our code implementation of the grammar includes the original license statement from SCAN.
    \item Our use of the permutation equivariant model from \citet{gordon2019permutation} uses the original implementation. The code is released under an MIT license. All source code files redistributed with our work contain the original license statement, as per the license requirement.
    \item The Transformer \citep{vaswani2017attention} and the CNN \citep{gehring2017convolutional} is implemented by the authors in PyTorch \citep{paszke2019pytorch} and do not use existing code implementations. However, the CNN was largely based on the implementation found in Fairseq \citep{ott2019fairseq}, which is released under an MIT license. We do not redistribute any of the original code files.
\end{enumerate}

\section{The use of AI assistants}
\label{appendix:ai_assistants}
The authors used AI assistants for spellchecking and sentence-level editing in the writing of the current manuscript. 

\end{document}

%% file: sections/introduction.tex
Human language is characterized by its combinatorial properties in syntax and semantics \citep{hadley1994systematicity}. For example, when someone knows what it means for Person A to see Person B, they can readily generalize to what it means for Person B to see Person A, without having been exposed to examples where Person B is used in the subject position \citep{fodor1988connectionism}. This type of bidirectional systematic generalization has long been considered a key feature of human language understanding \citep{chomsky1957syn}, and is closely related to the compositionality of natural language semantics, a contentious but widely studied topic in both formal semantics \citep{van1991logic} and NLP \citep{mccurdy-etal-2024-toward}.
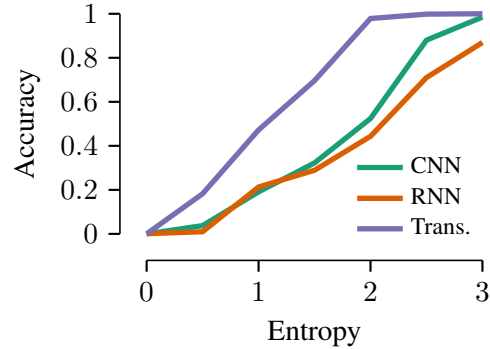
\begin{figure}
    \null\hfill%
    \input{figures/front_page_figure}%
    \hfill\hfill\null%
    \caption{Systematic generalization increases with the entropy of the training data for popular sequence-to-sequence models, even without built-in priors.}
    \label{fig:front_page}
\end{figure}

Early critics of connectionism, such as \citet{fodor1988connectionism}, argued that systematic generalization was incompatible with continuous representations. More than thirty years since this critique was initially presented, researchers still agree that achieving robust compositional behavior in neural networks remains an open problem, as shown by a recent survey by \citet{mccurdy-etal-2024-toward}. Research continues to demonstrate that current models struggle with systematic generalization \citep{lake2018generalization, hupkes2020compositionality, dziri2024faith, wold-etal-2024-compositional, gsmSymbolic2024}.

Some of the limitations of current models can be alleviated by implementing compositional priors, either in the architecture \citep{gordon2019permutation} or through the training procedure \citep{lake2023human}, but it remains unclear whether architectural constraints cause the limitations or whether they arise due to distributional properties of the data used for training.%
\begin{figure*}
	\input{figures/distrib_exp1.tex}
	\caption{A schematic overview of the two approaches to increasing the entropy level. Top: Vertical scaling by distribution mixing. Bottom: Horizontal scaling by incrementing the support of the distribution.}
    \label{fig:approach}
\end{figure*}
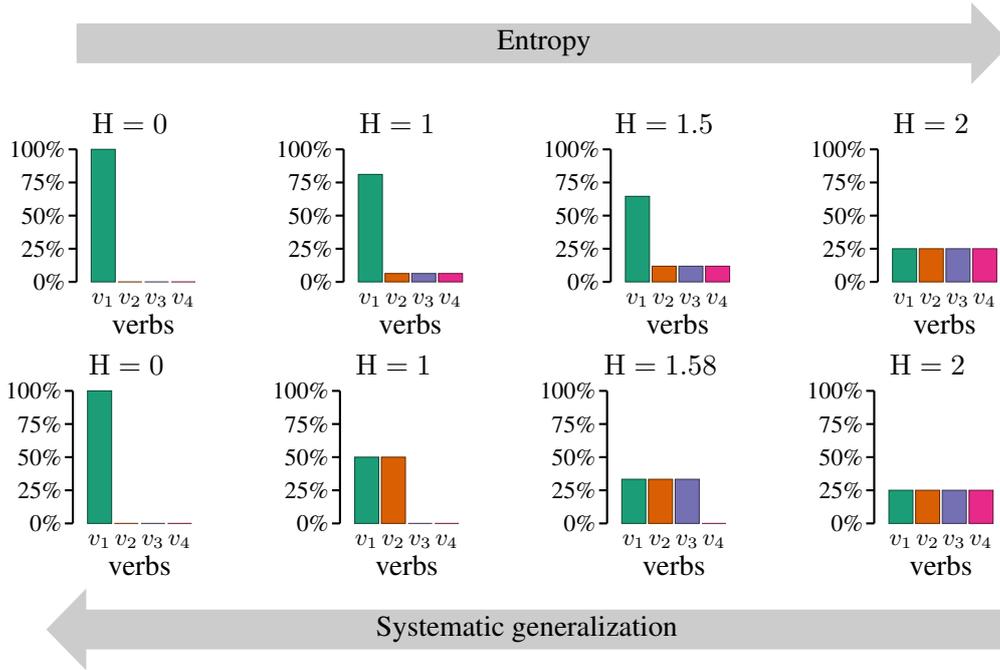

In this paper, we ask the following question: Can neural networks display systematic out-of-distribution generalization without \textit{any} built-in priors? To answer this, we take a data-centric approach. Specifically, we show how the degree of systematic generalization of common architectures relates to the information entropy of the distribution of component parts found in the training data.

Contrary to previous work, we find that popular sequence-to-sequence models used in NLP are capable of systematic generalization, even without built-in architectural priors. However, this is only the case when the training dataset contains high entropy. On a series of sequence-to-sequence datasets based on a modified version of the SCAN \citep{lake2018generalization} grammar, we show that model performance scales with the entropy of the training dataset, as illustrated in \cref{fig:front_page}, thereby creating a link between systematic generalization and information efficiency. 

 In summary, our contributions are: \emph{(i)} We develop and formalize a framework for studying systematic generalization based on information entropy; \emph{(ii)} We experiment with two different approaches to increasing entropy, and we compare the performance of four different model architectures for these two settings; and \emph{(iii)} we show that neural networks achieve systematic out-of-distribution generalization when the entropy of the training distribution is high, independently of the number of unique samples seen during training.

%% file: figures/front_page_figure.tex
\begin{tikzpicture}
    \begin{axis}[
        modern,
        width=60mm,
        height=45mm,
        xlabel={Entropy},
        ylabel={Accuracy},
        legend pos=south east,
        legend cell align={left},
        legend style={font=\small, draw=none, fill=none, at={(0.80,0.40)}, anchor=north},
        xmin=0,
        xmax=3,
        ymin=0,
        ymax=1,
        clip=false
    ]
    \addplot[
        dark1,
        line width=2pt,
        mark=none,
    ] table[x=entropies, y=accuracy, y] {figures/data/experiment1_cnn.dat};
    \addlegendentry{CNN}   
    \addplot[
        dark2,
        line width=2pt,
        mark=none,
    ] table[x=entropies, y=accuracy, y] {figures/data/experiment1_rnn.dat};
    \addlegendentry{RNN}   
    \addplot[
        dark3,
        line width=2pt,
        mark=none,
    ] table[x=entropies, y=accuracy, y] {figures/data/experiment1_transformer.dat};
    \addlegendentry{Trans.}   
    \end{axis}
\end{tikzpicture}

%% file: figures/distrib_exp1.tex
\centering
\begin{tikzpicture}
    \verbdistribution{}{1, 0, 0, 0}{(0pt, 0pt)}
	\verbdistribution{}{0.8107105, 0.0630965, 0.0630965, 0.0630965}{(100pt, 0pt)}
	\verbdistribution{}{0.645457, 0.118181, 0.118181, 0.118181}{(200pt, 0pt)}
	\verbdistribution{}{0.25, 0.25, 0.25, 0.25}{(300pt, 0pt)}
	\node at ( 20pt, 10pt+\probabilityheight) {\(\mathrm{H}=0\)};
	\node at (120pt, 10pt+\probabilityheight) {\(\mathrm{H}=1\)};
	\node at (220pt, 10pt+\probabilityheight) {\(\mathrm{H}=1.5\)};
	\node at (320pt, 10pt+\probabilityheight) {\(\mathrm{H}=2\)};
	
	\begin{scope}[shift={(0, \probabilityheight)}]
		\draw[line width=15pt, black!20, -{Triangle[length=15pt, width=30pt]}] (0, 40pt) -- +(350pt, 0);
		\node at (175pt, 40pt) {Entropy};

	\end{scope}
\end{tikzpicture}

\smallskip

\begin{tikzpicture}
    \verbdistribution{}{1, 0, 0, 0}{(0pt, 0pt)}
	\verbdistribution{}{0.5, 0.5, 0, 0}{(100pt, 0pt)}
	\verbdistribution{}{0.333, 0.333, 0.333, 0}{(200pt, 0pt)}
	\verbdistribution{}{0.25, 0.25, 0.25, 0.25}{(300pt, 0pt)}
	\node at ( 20pt, 10pt+\probabilityheight) {\(\mathrm{H}=0\)};
	\node at (120pt, 10pt+\probabilityheight) {\(\mathrm{H}=1\)};
	\node at (220pt, 10pt+\probabilityheight) {\(\mathrm{H}=1.58\)};
	\node at (320pt, 10pt+\probabilityheight) {\(\mathrm{H}=2\)};


	\draw[line width=15pt, black!20, {Triangle[length=15pt, width=30pt]}-] (-10pt, -40pt) -- +(360pt, 0);
	\node at (170pt, -40pt) {Systematic generalization};
\end{tikzpicture}

%% file: sections/background.tex
Systematic generalization is typically described as a robust mode of composition, where a system is able to combine parts of its training data in a way that enables generalization to novel combinations. As such, systematic generalization relates to compositionality, a topic which has both a long tradition and multiple formal definitions, most prominently in formal semantics \citep{van1991logic, pagin2010compositionality}. These definitions, however, often prove too restrictive when working with natural languages. As a result, researchers frequently introduce problem-specific relaxations, leading to ambiguity over time regarding the exact meaning of terms such as systematic generalization and compositionality.

Despite there being some ambiguity regarding the formal definitions, recent work shows that there is high agreement among researchers on the following, more informal definition of what constitutes compositional behavior:

\begin{definition}[CB]
    \label{def:cb}\upshape
    ``When a model receives an input \(I\) that humans conceive as composed of component parts, if the model produces correct outputs for those parts (in isolation or in other combinations), then it will also produce a correct output for \(I\)~\citep[p.~9324]{mccurdy-etal-2024-toward}.''
\end{definition}

This definition, however, does not consider the conditions from which a system learns to produce these outputs. Consequently, compositional behavior differs from compositional \emph{generalization}. While the former can be assessed through normal benchmarks, the latter requires information about the distribution the model generalizes from.

Although generalization is a loosely defined term in machine learning research, it generally refers to how well a model performs on a held-out dataset after being trained on data that has some distributional relationship with the held-out data \citep{hupkes_taxonomy_2023}. When training and test distributions are similar, we refer to this as \textit{in-distribution}, and when they differ, we typically refer to this as \textit{out-of-distribution}. Naturally, the classification of a generalization problem as in-distribution or out-of-distribution depends on which aspects of the distribution are considered. In this work, we focus on \textit{systematic} generalization as an out-of-distribution problem, where the differences in the distribution of component parts and their combinations between the training and test data are the relevant aspects. 

\subsection{Systematic Generalization}\label{section:sys_gen}
Following earlier works, we will refer to the systematic generalization abilities of a model as the \textit{systematicity} of that model. Our view of systematicity is based on \citet{hadley1994systematicity}, who in contrast to earlier work by \citet{fodor1988connectionism} distinguished between \textit{degrees} of systematicity. In the following paragraphs, we present and define these degrees under the assumption that we are considering a system tasked to generalize from a training corpus to a held-out test corpus:\footnote{\citet{hadley1994systematicity} also discusses an intermediate degree between weak and strong systematicity, so-called \textit{quasi-systematicity}, but we leave this out for brevity, as it does not relate directly to our use case.} 

\begin{definition}[Weak systematicity]
    \label{def:weak}\upshape
    A system displays \textit{weak systematicity} if the system requires that all component parts that occur in the training data also occur at every permissible syntactic position. Even if the model succeeds at a test set that contains novel samples, none of the components in these samples occur at a novel syntactic position, essentially making the problem in-distribution with respect to the systematic properties of the data.
\end{definition}


\begin{definition}[Strong systematicity]
    \label{def:strong}\upshape
    Now, a system displays \textit{strong systematicity} if it \emph{(i)} displays weak systematicity, meaning that it can generalize in-distribution, and \emph{(ii)} it can process novel embedded sentences containing previously learned component parts that occur in syntactic positions where they do not appear in the training data, neither in embedded sentences nor in non-embedded sentences, and \emph{(iii)} a significant fraction of the possible combinations of component parts and their permissible syntactic position should be left out, based on the observation that humans learn grammars based on largely incomplete data \citet{pinker1989learnability}. In this context, an embedded sentence can be thought of as a syntactic constituent in the overall sentence. For example, if a system can process the following novel sentence \textit{Mathilde knows Sigurd and Sigurd knows Olav}, which has the sentences \textit{Mathilde knows Sigurd} and \textit{Sigurd knows Olav} as embedded sentences, without having seen the name \textit{Mathilde} previously used in the subject position in a sentence with constituents nor in a sentence with no constituents, then it is strongly systematic w.r.t to how it can generalize to the novel use of the name \textit{Mathilde}. On the other hand, if the system can only process this novel sentence if it has seen \textit{Mathilde} being used in the same syntactic position, for example in the sentence \textit{Mathilde knows Olav}, then it is not strongly systematic. 
\end{definition}

\paragraph{Systematic generalization in embedded sentences.}
In this work, we focus on a relaxed version of strong systematicity as defined in \cref{def:strong}. We relax requirement \emph{(ii)} from the original definition so that we only work with samples that have embedded sentences. As we show in the following section, this makes it easier to quantify the level of systematicity required for a given generalization problem, which is the main focus of this work. Furthermore, we also relax \emph{(ii)} so that it allows component parts of the test set to occur at the same syntactic position in \textit{some} embedded sentence in the train set, but not in an embedded sentence that occurs at the same position in the overall sequence. 

In order to compare models with and without compositional priors from the literature, we base our experiments on the SCAN grammar \citep{lake2018generalization}. This grammar has few permissible syntactic positions for each part of speech, making the original requirement from \citet{hadley1994systematicity} too restrictive with respect to the overall sample size needed by neural networks. In the following section, we formalize our framework and elaborate on these choices. 

%% file: sections/measuring_systematicity.tex
In this work, we demonstrate that sequence-to-sequence models can generalize to embedded sentences where component parts occur at novel syntactic positions in the overall sequence. For our experiments, we construct datasets where each dataset sample contains an input sequence $x \in X$ and an output sequence $y \in Y$, based on a modified version of the SCAN grammar. We discuss the specifics of this dataset generation procedure, including details about the context-free grammar that generates $x$, in \cref{section:data_generation}. In this section, we define a framework for measuring the \textit{degree} of systematicity displayed by a systematic generalization problem. 

\subsection{Task Description}
Throughout this paper, $x$ comprises two sentences conjoined by a conjunctive phrase.\footnote{In contrast to the original SCAN, we do not have sentences without a conjunctive phrase.} Consequently, $x$ can be modeled as a triple, with a conjunction $\operatorname{c} \in C$, and the two embedded sentences $e_1, e_2\in\Sigma^*$ as arguments, $x=(e_1, \operatorname{c}, e_2)$. The component parts of both embedded sentences are drawn from the same vocabulary $\Sigma = V \cup A$, where $V$ is a set of verbs and $A$ is a set of adverbial phrases, e.g.: \textsc{jump, run} $\in V$, \textsc{twice} $\in A$, and \textsc{and} $\in C$. There exists a deterministic procedure $\varphi$ for transforming $x$ into~$y$:

\begin{trivlist}\item
    $\varphi(\textsc{run twice and jump}) = \textit{RUN RUN JUMP}.$
\end{trivlist}

The goal of the described task is to learn a function $f$ that approximates $\varphi$, which requires modeling the conditional distribution of output sequences given input sequences in the training data, i.e.~$p^{\text{train}}(y \mid x)$. See \cref{tab:scan_examples} for more examples generated by our modified SCAN grammar. 

\subsection{Component Distributions}

Since we are interested in systematic generalization, we also need to define the distribution of component parts in $e_1$ and $e_2$, and we are specifically interested in the case where the distribution of $e_1$ and $e_2$ is different in the training and test data, but they share a similar distribution when marginalized over position, which in this case refers to which side of the conjunctive phrase they appear in.

In SCAN (\cref{appendix:scan_grammar}), $e_1$ and $e_2$ always contain \textbf{one} verb in $V$ and zero or more adverbial phrases from $A$ that control how many times that verb should be repeated, and in what direction it is executed. To create a systematic generalization problem that adheres to the definition given in \cref{section:sys_gen}, it is necessary to set the support of $e_1$ and $e_2$ to be different in the training and test distributions.  For simplicity, we limit our study to focus on the setting where the support of the verbs $V$ is different between the two distributions. This means that we can create a generalization problem by restricting a verb $v_1$ to never occur in $e_1$ but to occur in $e_2$ for $p^{train}$, as illustrated by \cref{fig:example_verb_distributions}.

For $p^{\text{test}}$ we invert the distribution of $e_1$, making it a degenerate distribution with single-point support for the verb that is not in $e_1$ for $p^{\text{train}}$, while $e_2$ has the same distribution as in $p^{\text{train}}$. When generalizing from $p^{\text{train}}$ to $p^{\text{test}}$ in this scenario, a model will be exposed to all verbs in all permissible syntactic positions in \textit{some} embedded sentence, but $v_1$ will never be seen in $e_1$ during training, while $e_1$ in $p^{\text{test}}$ always contains $v_1$.

\begin{figure}
    \centering
    \input{figures/e2_verb_uniform}
    \caption{Example of distributions of verbs for $e_1$ (left) and $e_2$ (right).}
    \label{fig:example_verb_distributions}
\end{figure}
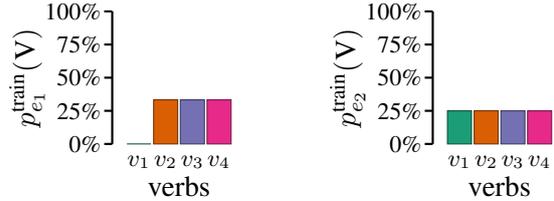
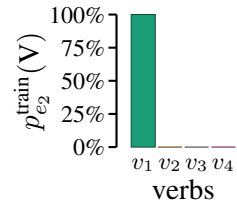
\begin{figure}[ht]
    \centering
    \input{figures/e2_verb_degenerate}
    \caption{Probabilities in the degenerate case for $e_2$}
    \label{fig:verb_degenerate}
\end{figure}

\subsection{Systematicity as Entropy}
The main contribution of this work is to demonstrate that the difficulty of generalizing from $p^{\text{train}}$ to $p^{\text{test}}$ in the scenario described above depends on the \textit{entropy} $\mathrm{H}$ of the distribution of verbs in $e_2$ for $p^{\text{train}}$:
\begin{equation*}
    \mathrm{H}^{\text{train}}_{e_2}(\textrm{V}) =  - \sum_{v \in V} p^{\text{train}}_{e_2}(v) \log_2 p^{\text{train}}_{e_2}(v),
\end{equation*}
where $\textrm{V}$ is the verb random variable and $V$ is the set of all verbs. For brevity, we will refer to $\mathrm{H}^{\text{train}}_{e_2}(\textrm{V})$ as $\mathrm{H}$. When the verbs in $e_2$ have a degenerate distribution, as shown in \cref{fig:verb_degenerate}, the entropy is zero, and when $e_2$ is distributed uniformly, as shown in \cref{fig:example_verb_distributions}, the entropy is at its maximum. Decreasing $\mathrm{H}$ within this interval increases the difficulty of the systematic generalization problem. Note that even at maximum entropy, the generalization problem is still out-of-distribution, as the restricted verb is never seen in $e_1$. In \cref{section:experiments}, we show empirically that the performance of neural network architectures scales with $\mathrm{H}$ on this type of generalization problem.

\paragraph{The degenerate case.}
We note that $\mathrm{H}=0$ constitutes a special case in the described framework. If $e_2$ is described by a degenerate distribution, then instances of $e_2$ will always contain the same $v_1 \in V$. Now, if $\vert C \vert = 1$, there is no way of decoupling the semantics of the conjunctive phrase and $v_1$, as these will always appear together in $X$\@. In this case, we argue, no system will be able to generalize from $p^{\text{train}}$ to $p^{\text{test}}$. Consequently, it is necessary to set $\vert C \vert > 1$ for evaluating performance at $\mathrm{H}=0$. 
\begin{table*}
    \centering
    \begin{tabular}{ll}
        \toprule
        Input sequence $x$ & Output sequence $y$ \\
        \midrule
        \textcolor{dark1}{squat opposite right} and \textcolor{dark2}{squat} &  RTURN RTURN SQUAT SQUAT \\
        \textcolor{dark1}{squat twice} and \textcolor{dark2}{crawl opposite left} & SQUAT SQUAT LTURN LTURN CRAWL \\
        \textcolor{dark1}{sprint right twice} after \textcolor{dark2}{sprint left} & LTURN SPRINT RTURN SPRINT RTURN SPRINT \\
        \textcolor{dark1}{lunge opposite left} and \textcolor{dark2}{look thrice} & LTURN LTURN LUNGE LOOK LOOK LOOK \\
        \bottomrule
    \end{tabular}
    \caption{Examples from our generated datasets, with \textcolor{dark1}{green} indicating $e_1$ and \textcolor{dark2}{orange} indicating $e_2$.}
    \label{tab:scan_examples}
\end{table*}
\subsection{Increasing Entropy}
\label{sec:increasing_entropy}

In this work, we experiment with two different approaches to increasing $H$, as illustrated in \cref{fig:approach}. In the following sections, we formalize these approaches, and in \cref{section:experiments} we demonstrate how the performance of neural architectures scales for each type of increment. 

\subsubsection{Distribution Mixing}
\label{sec:mixing}

Let $\mathcal{U}$ be a uniform distribution with support over $V \setminus \{v_1\}$, and let $\mathcal{D}$ be a degenerate distribution over $V$ with single-point support for $v_1$. Our first approach consists in defining a family of distributions parametrized by $\lambda$ mixing $\mathcal{U}$ and $\mathcal{D}$:
\begin{equation*}
    p_{e_2}^{\text{train}}(\mathrm{V)} = \lambda \mathcal{U} + (1-\lambda) \mathcal{D},
\end{equation*}
Then, the entropy $\mathrm{H}$ is at the minimum when $\lambda=0$ and at the maximum when $\lambda = 1 - 1/|V|$, which is equal to a uniform distribution over $V$. Consequently, increasing $\mathrm{H}$ can be achieved by tending $\lambda$ towards $1 - 1/|V|$. This approach to increasing $\mathrm{H}$ is also illustrated in the top part of \cref{fig:approach}.

\subsubsection{Incremental Support}
\label{sec:inc_supp}

In our second approach, we define another family of distributions, all uniforms over increasingly larger supports \(S_i\subseteq V\):
\begin{equation*}
    \{v_1\} = S_1 \subset S_2 \subset \cdots \subset S_{|V|} = V
\end{equation*}
The uniform distribution over the smallest support \(S_1\) is equivalent to the degenerate distribution \(\mathcal{D}\).
As \(i\) increases, \(\mathrm{H}\) increases logarithmically until reaching maximum entropy at \(i=|V|\) where we have the uniform distribution over \(V\).
This approach to increasing $\mathrm{H}$ is also illustrated in the bottom part of \cref{fig:approach}.





\subsection{Data Generation} \label{section:data_generation}
As posited in the previous sections, the quantification of systematic generalization requires information about the distribution of parts and combinations observed during training. This level of granularity is generally intractable when dealing with real-world corpora. Without this information, however, it is difficult to control for statistical patterns that permit a non-compositional solution. Studies on systematicity typically overcome this problem by studying synthetic languages where the distribution can be known a priori. This is also the approach taken in this work.

To generate datasets samples $(x, y) \in X \times Y$ we define a context-free grammar (CFG) based on the SCAN grammar \citep{lake2018generalization}. An overview of the complete vocabulary can be found in \cref{appendix:scan_grammar}, while \cref{tab:scan_examples} shows examples from our generated data. Importantly, we set $\vert V \vert$ to 8, as opposed to the original four verbs, thereby increasing the number of possible sequences generated by the grammar. We also remove the \texttt{turn} operators, as they are more syntactically constrained than other verbs, as was already done in \citet{gordon2019permutation}. As previously stated, all instances of $x$ contain embedded sentences $e_1$ and $e_2$.

In our experiments, we use the standard conjunctive phrases in SCAN $\vert C \vert = 2$. Furthermore, we let the two conjunctive phrases take the embedded sentences in the opposite order: if $v_1$ always occurs in $e_2$ for $c_1$, then $v_1$ will always occur in $e_1$ for $c_2$.

%% file: figures/e2_verb_uniform.tex
\begin{tikzpicture}
	\verbdistribution{\(p_{e_1}^{\text{train}}(\textrm{V})\)}{0, 0.333, 0.333, 0.333}{(0pt, 0pt)}
	\verbdistribution{\(p_{e_2}^{\text{train}}(\textrm{V})\)}{0.25, 0.25, 0.25, 0.25}{(120pt, 0pt)}
\end{tikzpicture}

%% file: figures/e2_verb_degenerate.tex
\begin{tikzpicture}
\verbdistribution{\(p_{e_2}^{\text{train}}(\textrm{V})\)}{1.0, 0.0, 0.0, 0.0}{(80pt, 0pt)}
\end{tikzpicture}

%% file: sections/experiments.tex
We validate our framework by evaluating the three most common sequence-to-sequence architectures: the Transformer, the RNN, and the CNN. We also run a model with built-in compositional priors to showcase how a model that encodes verb equivariance can solve the task even when $\mathrm{H}$ is low.
We have two main experimental settings, corresponding to the two methods for increasing $\mathrm{H}$ described in \cref{sec:mixing} and \cref{sec:inc_supp}. We also conduct a supporting experiment to control for a potential confounding effect resulting from an increased sample size, and an experiment that estimates the influence of different position encoding types. For all experimental settings, $\mathrm{H}=0$ constitutes the most difficult generalization problem, and $\mathrm{H}= \log_2 \vert V \vert = 3$ constitutes the easiest problem. We average and report results from five differently seeded runs. The following sections provide details about our experimental setup.\footnote{Data generation scripts and code to reproduce our experiments can be found at \url{https://github.com/ltgoslo/systematicity-entropy}.}

\subsection{Models}
\paragraph{RNN.} For our RNN we use a similar configuration as previous work on SCAN \citep{lake2018generalization, gordon2019permutation}: a bidirectional encoder-decoder with attention. Details on the implementation and choice of hyperparameters can be found in \cref{appendix:model_details:rnn}.

\paragraph{CNN.} For our CNN we use the encoder-decoder architecture from \citet{gehring2017convolutional}, which has been proven effective on SCAN in previous work \citep{dessi-baroni-2019-cnns}. As in the RNN, this CNN uses an attention mechanism between the encoder and decoder. Hyperparameters can be found in \cref{appendix:model_details:cnn}.

\paragraph{Transformer.} We use the original encoder-decoder Transformer from \citet{vaswani2017attention}. We use this for two reasons. Firstly, this makes comparison with previous work on systematicity easier, as this has been used in previous work such as \citet{hupkes2020compositionality} and \citet{lake2023human}. Secondly, recent work on machine translation has shown that encoder-decoders outperform the more widely used decoder-only formulation of the Transformer on similar parameter sizes \citep{pitorro-etal-2024-effective}. As SCAN is framed and modeled as a sequence-to-sequence task, the encoder-decoder formulation is the most reasonable choice. We use absolute position encodings for the two main experiments, as well as Gated Linear Units as the activation function \citep{shazeer2020glu}. Details on the implementation and choice of hyperparameters can be found in \cref{appendix:model_details:transformer}. 

\paragraph{Permutation-equivariant model.} Lastly, we experiment with a model that has built-in architectural priors. We use the encoder-decoder from \citet{gordon2019permutation}, which enforces the verb equivariance found in SCAN using a cyclic group. This group is used in the encoding of the input sequences in a way that pools the representation of any verb with the representation of the other verbs, even when they are not present in the input. This is a strong structural prior that enables the model to solve the original SCAN splits efficiently. However, this approach is dataset-dependent, as the elements of the cyclic group must be defined manually. We use this model primarily as a sanity check for our experiments, but also as a comparison for the other architectures, comparing their efficiency to an idealized case. 

\subsection{Experiment 1: Vertical scaling}\label{sec:experiment_1}
For Experiment 1, $p^{\text{test}}$ has a degenerate distribution for $e_1$, where $v_1 \in V$ always occur, while $e_2$ has a uniform distribution over $V$. For $p^{\text{train}}$, $e_1$ has a uniform distribution over $V \setminus \{v_1\}$, and $e_2$ is a mixture of a degenerate distribution at $v_1$ and a uniform distribution over $V \setminus \{v_1\}$, where we increase the total samples drawn from the uniform while decreasing the number drawn from the degenerate distribution. By doing this, we increase $\mathrm{H}$ as described in section \cref{sec:mixing}. We refer to this increase in $\mathrm{H}$ as \textit{vertical scaling}, as the distribution over the verbs increases towards the uniform. The models are trained on 6\,000 unique samples and evaluated on 7\,056 samples for all values of $\mathrm{H}$. 

\begin{figure*}
\begin{subfigure}{.5\textwidth}
  \input{figures/experiment1}
  \caption{Experiment 1: Vertical scaling of entropy}
  \label{fig:exp1:vertical}
\end{subfigure}%
\begin{subfigure}{.5\textwidth}
  \input{figures/experiment2}
  \caption{Experiment 2: Horizontal scaling of entropy}
  \label{fig:exp2:horizonal}
\end{subfigure}
\caption{Results from the Experiments~1 \& 2. The accuracy and standard deviations are from five seeds.}
\label{fig:main_experiments}
\end{figure*}
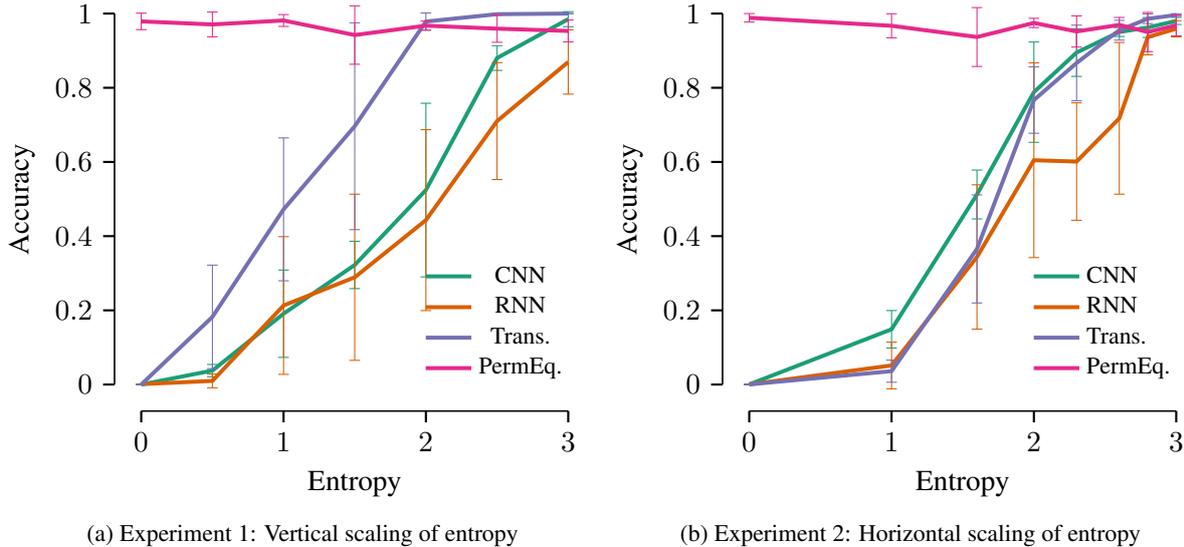  
\subsection{Experiment 2: Horizontal Scaling}\label{sec:experiment_2}
For Experiment 2, $p^{\text{test}}$ is the same as for Experiment 1. For $p^{\text{train}}$, $e_1$ is also the same as for Experiment 1, but $e_2$ is a uniform distribution over an increasing number of verbs, starting with $\{v_1\}$ and ending at $V$, as described in \cref{sec:inc_supp}. We refer to this increase in $\mathrm{H}$ as \textit{horizontal scaling}, as the increase of $\mathrm{H}$ is achieved by adding another verb to the support of $e_2$. The models are trained on all grammatical combinations in $\Sigma^*$ permitted by the support at each level of $\mathrm{H}$. When $\mathrm{H}=2$, for example, the support of the distribution of $e_2$ contains four verbs, and the training set contains all possible pairs $(x, y) \in X \times Y$ under these constraints.

To ensure a fair comparison across entropy levels, we train all models for the same number of optimization steps, so for $\mathrm{H}=1$, which has fewer training samples than $\mathrm{H}=3$, we do more iterations over the full data.

\subsection{Effect of Number of Unique Samples}
\label{sec:experiment_unique_samples}
As $\mathrm{H}$ describes the relative distribution of verbs in $e_2$, it is possible to construct datasets of varying sizes that have the same $\mathrm{H}$. This can be achieved by increasing the total volume of unique samples while maintaining the same relative distribution. To verify that performance scales with $\mathrm{H}$ independently of the number of unique samples seen during training, we create datasets that have the same entropy as in the previous experiments, but where we vary the number of total unique samples. We experiment with three different sample sizes: 3\,000, 4\,000, and 6\,000, using the vertical scaling setting. All models run for the same number of optimization steps for all sample sizes. 

\subsection{Effect of Positional Embedding Type}
\label{sec:experiment_position_embedding}

In our experiments, there is an intuitive connection between the generalization from $p^{\text{train}}$ to $p^{\text{test}}$ and the ability to encode verb equivariance. This is exemplified by \citet{gordon2019permutation}, which achieves high performance on the original SCAN tasks precisely by enforcing verb equivariance in the model architecture explicitly. 

As the verbs always occur on the first token position of $e_1$ and $e_2$, positional information is central to our experiments. To estimate the effect of positional information, we show how the choice of positional encoding scheme affects performance in the vertical scaling setting. In addition to the original absolute position embeddings from \citet{vaswani2017attention}, we experiment with embeddings that encode relative positional information through disentangled attention \citep{he2021deberta} and rotary position embeddings (RoPE; \citealp{su2024roformer}). In this experiment, we focus on the Transformer. Details on hyperparameter selection can be found in \cref{appendix:model_details:positional}

\subsection{Results and Discussion}
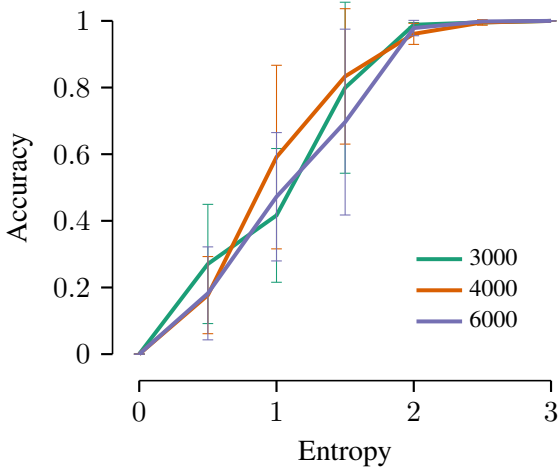
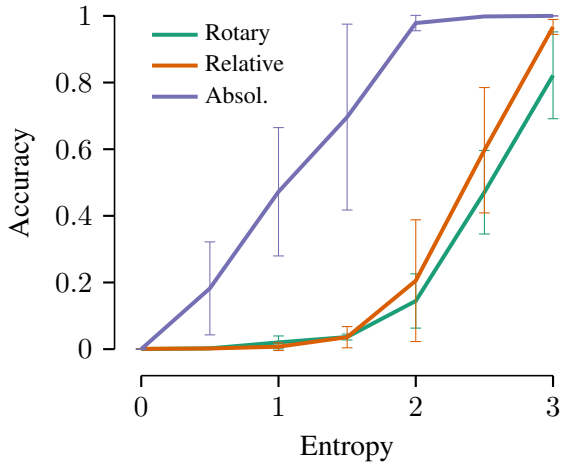
\begin{figure*}
\begin{subfigure}{.5\textwidth}
  \input{figures/experiment_sample_effect}
    \caption{Effect of unique sample size on the Transformer}
    \label{fig:unique_samples}
\end{subfigure}%
\begin{subfigure}{.5\textwidth}
  \input{figures/experiment_embedding}
    \caption{Effect of positional embedding type for the Transformer}
    \label{fig:embedding}
\end{subfigure}
\caption{The result of the supplemental experiments. The accuracy and standard deviations are from five seeds.}
\label{fig:supporting_experiments}
\end{figure*}
\label{sec:results}
The results of Experiment 1 and Experiment 2 can be found in \cref{fig:main_experiments}. We observe a clear positive relationship between performance and $\mathrm{H}$ for both experiments, demonstrating that systematic generalization scales with entropy. Unsurprisingly, the permutation-equivariant model, which has a strict compositional prior, solves both tasks at all entropy levels. This shows that the task is solvable even at $\mathrm{H}=0$ when enforcing such a prior, but that vanilla architectures generally require much higher entropies for this type of generalization.

In Experiment 1, the Transformer outperforms both the RNN and CNN across all levels of $\mathrm{H}$. In Experiment 2, the performances of all models are statistically similar, falling within one standard deviation of each other, but with the CNN performing the best on average in the reported runs. 

We also observe that for the Transformer, a reduction in the support of the distribution is more detrimental to the performance than a step towards the degenerate distribution: At $\mathrm{H}=2$, the Transformer has close to 100\% accuracy for Experiment 1, but for the same level of $\mathrm{H}$ in Experiment 2, it achieves around 80\% accuracy. This is not the case for the RNN and CNN, where we observe the opposite: The increase in support that increases $\mathrm{H}$ from 1 to 2 in Experiment 2 gives a higher performance increase than what the same increase in $\mathrm{H}$ does for Experiment 1. 

We argue that these results indicate that the Transformer is more information-efficient than the RNN and CNN under full support, but that the models are similar under lacking support. In practice, this means that the Transformer is quicker to generalize to novel uses of component parts, as long as the overall syntactic pattern---constituted by the conjunctive phrase in our experiments---is frequent in the training data. This interpretation aligns with the findings of \citet{loula-etal-2018-rearranging}, showcasing how our framework can provide a theoretical explanation for previous results.

We note that even though all models can generalize under high entropy for both experiments, their performance is subpar for low entropy, e.g.~$\mathrm{H}=1$. This indicates a rather high lower bound on the entropy required for systematic generalization, both for vertical and horizontal scaling, and it shows that all models are unable to meet requirement \emph{iii} from \cref{def:strong}. Furthermore, our results show that systematic generalization remains an open problem when there is no prior, and it remains unclear whether or not it is possible to build a model that performs well on low entropy without architectural priors as in \citet{gordon2019permutation} or priors in the training procedure as in \citet{lake2023human}.

\subsubsection{Supplemental results}
In \cref{fig:unique_samples}, we show the results of varying the number of unique samples at each level of $\mathrm{H}$. We observe that the positive relationship between performance and $\mathrm{H}$ is independent of the number of unique samples. The Transformer performs similarly when trained on 3\,000 to 6\,000 unique samples. During hyperparameter selection (\cref{appendix:model_details}), we also observe that performance does not scale with model size. This indicates that achieving higher accuracy on lower levels of $\mathrm{H}$ is not achievable by scaling the number of unique data samples or model size for the same architecture.

In \cref{fig:embedding}, we report the performance of different positional encoding types. Although all encoding types achieve near-ceiling performance at high entropy, absolute encoding performs better for all levels of $\mathrm{H}$. From our task formulation in \cref{section:measuring}, it is clear that positional information is crucial for solving the tasks. We attribute the efficiency of absolute encodings to the fact that the positional information of the verbs is independent of the remainder of the sequence. This invariance has to be learned by the other embedding types, making the absolute variant more information-efficient. 

%% file: figures/experiment1.tex
\begin{tikzpicture}
    \begin{axis}[
        modern,
        width=72mm,
        height=65mm,
        xlabel={Entropy},
        ylabel={Accuracy},
        grid=minor,
        legend style={font=\small, draw=none, fill=none, at={(0.83,0.35)}, anchor=north},
        xmin=0,
        xmax=3,
        ymin=0,
        ymax=1,
        clip=false,
        ytick distance=0.2
    ]
    
    \addplot[
        dark1,
        line width=1.5pt,
        error bars/.cd,
        y dir=both,
        y explicit,
    ] table[x=entropies, y=accuracy, y error=std] {figures/data/experiment1_cnn.dat};
    \addlegendentry{CNN}
    
    \addplot[
        dark2,
        line width=1.5pt,
        error bars/.cd,
        y dir=both,
        y explicit
    ] table[x=entropies, y=accuracy, y error=std] {figures/data/experiment1_rnn.dat};
    \addlegendentry{RNN}

    \addplot[
        dark3,
        line width=1.5pt,
        error bars/.cd,
        y dir=both,
        y explicit
    ] table[x=entropies, y=accuracy, y error=std] {figures/data/experiment1_transformer.dat};
    \addlegendentry{Trans.}

    \addplot[
        dark4,
        line width=1.5pt,
        error bars/.cd,
        y dir=both,
        y explicit
    ] table[x=entropies, y=accuracy, y error=std] {figures/data/experiment1_permutation.dat};
    \addlegendentry{PermEq.}
    
    \end{axis}
    \end{tikzpicture}

%% file: figures/experiment2.tex
\begin{tikzpicture}
    \begin{axis}[
        modern,
        width=72mm,
        height=65mm,
        xlabel={Entropy},
        ylabel={Accuracy},
        grid=minor,
        legend pos=south east,
        legend cell align={left},
        legend style={font=\small, draw=none, fill=none, at={(0.83,0.35)}, anchor=north},
        xmin=0,
        xmax=3,
        ymin=0,
        ymax=1,
        clip=false,
        ytick distance=0.2
    ]
    \addplot[
        dark1,
        line width=1.5pt,
        error bars/.cd,
        y dir=both,
        y explicit
    ] table[x=entropies, y=accuracy, y error=std] {figures/data/experiment2_cnn.dat};
    \addlegendentry{CNN}
    
    \addplot[
        dark2,
        line width=1.5pt,
        error bars/.cd,
        y dir=both,
        y explicit
    ] table[x=entropies, y=accuracy, y error=std] {figures/data/experiment2_rnn.dat};
    \addlegendentry{RNN}

    \addplot[
        dark3,
        line width=1.5pt,
        error bars/.cd,
        y dir=both,
        y explicit
    ] table[x=entropies, y=accuracy, y error=std] {figures/data/experiment2_transformer.dat};
    \addlegendentry{Trans.}

    \addplot[
        dark4,
        line width=1.5pt,
        error bars/.cd,
        y dir=both,
        y explicit
    ] table[x=entropies, y=accuracy, y error=std] {figures/data/experiment2_permutation.dat};
    \addlegendentry{PermEq.}
    
    \end{axis}
    \end{tikzpicture}

%% file: figures/experiment_sample_effect.tex
\begin{tikzpicture}
    \begin{axis}[
        modern,
        width=70mm,
        height=60mm,
        xlabel={Entropy},
        ylabel={Accuracy},
        grid=minor,
        legend pos=south east,
        legend cell align={left},
        legend style={font=\small, draw=none, fill=none, at={(0.80,0.35)}, anchor=north},
        xmin=0,
        xmax=3,
        ymin=0,
        ymax=1,
        clip=false,
        ytick distance=0.2
    ]
    
    \addplot[
        dark1,
        line width=1.5pt,
        error bars/.cd,
        y dir=both,
        y explicit
    ] table[x=entropies, y=accuracy, y error=std] {figures/data/sample_low.dat};
    \addlegendentry{3000}
    
    \addplot[
        dark2,
        line width=1.5pt,
        error bars/.cd,
        y dir=both,
        y explicit
    ] table[x=entropies, y=accuracy, y error=std] {figures/data/sample_medium.dat};
    \addlegendentry{4000}

    \addplot[
        dark3,
        line width=1.5pt,
        error bars/.cd,
        y dir=both,
        y explicit
    ] table[x=entropies, y=accuracy, y error=std] {figures/data/sample_high.dat};
    \addlegendentry{6000}

    \end{axis}
    \end{tikzpicture}

%% file: figures/experiment_embedding.tex
\begin{tikzpicture}
    \begin{axis}[
        modern,
        width=70mm,
        height=60mm,
        xlabel={Entropy},
        ylabel={Accuracy},
        grid=minor,
        legend pos=north west,
        legend cell align={left},
        legend style={font=\small, draw=none, fill=none, at={(0.00,0.85)}, anchor=west},
        xmin=0,
        xmax=3,
        ymin=0,
        ymax=1,
        clip=false,
        tick align=outside,
        ytick distance=0.2
    ]
    
    \addplot[
        dark1,
        line width=1.5pt,
        error bars/.cd,
        y dir=both,
        y explicit
    ] table[x=entropies, y=accuracy, y error=std] {figures/data/embedding_rope.dat};
    \addlegendentry{Rotary}
    
    \addplot[
        dark2,
        line width=1.5pt,
        error bars/.cd,
        y dir=both,
        y explicit
    ] table[x=entropies, y=accuracy, y error=std] {figures/data/embedding_rel.dat};
    \addlegendentry{Relative}

    \addplot[
        dark3,
        line width=1.5pt,
        error bars/.cd,
        y dir=both,
        y explicit
    ] table[x=entropies, y=accuracy, y error=std] {figures/data/experiment1_transformer.dat};
    \addlegendentry{Absol.}

    \end{axis}
    \end{tikzpicture}

%% file: sections/previous_work.tex
In this work, we base our experiments on a version of the SCAN grammar from \citet{lake2018generalization}. The original grammar has been studied in other works, such as \citet{loula-etal-2018-rearranging, dessi-baroni-2019-cnns, gordon2019permutation}, and there has also been work on the related PCFG SET dataset from \citet{hupkes2020compositionality}.

The work most similar to ours is \citet{zhou-etal-2023-data}, who also studies how data-centric aspects affect compositional generalization. In our work, we see compositionality as characterized by a set of primitives composed by a composition function, and we focus on how models capture this composition function depending on properties of the primitives. Their work, however, does not make this distinction, as their complexification process modify both the composition function and the distribution of primitives. The main distinction is hence that our work relates compositionality to dataset-level regularities, while they relate it to sample-level features.

Our work is also similar to \citet{keysers2020measuring}, who quantifies the level of systematic generalization using distributional information. Specifically, they quantify the compositionality of an experiment as the divergence between the distribution of component parts and combinations in the training and test data. Their work differs from ours since they focus on the case where the distribution of component parts is as similar as possible between the training and test data, while our work focuses on the case where the distribution of component parts is inverted from the training to the test set for some parts of the input.

In another domain, \citet{Wiedermer2024first} approaches systematicity in an image generation problem by factorizing the latent variables of the input image. The authors show that the support of these variables must be the same in the training and test distributions for any model to generalize systematically. We take a more fine-grained approach and our results demonstrate that support alone is insufficient as a predictor of systematic generalization; the performance also depends on the entropy, and higher levels of entropy in the training distribution can compensate for lower support.

There is also recent work on formalizing compositionality in terms of data complexity \citep{elmoznino2024complexity}, kernel-theory \citep{lippl2025does}, using graph formalisms \citep{ram2024makes}, and decomposition strategies \citep{jarvis2023on}. Our work is similar to these in the sense that we try to formalize some properties of systematic generalization, which is closely tied to compositionality.

%% file: sections/conclusion.tex
In this work, we show that systematic generalization of sequence-to-sequence models scales with information entropy. Through supporting experiments, we also show that this relationship is independent of the number of unique samples seen during training. Contrary to previous work, our findings demonstrate that these models are capable of systematic generalization even without any built-in architectural priors that incentivize compositional solutions. However, this requires that the training data have high entropy w.r.t.~the parts that the target composition function operates on. This poses a new question: Is it possible to achieve systematicity also at low entropy \textit{without} compositional priors? Through our formalizations, we hope to facilitate future work that attempts to answer this question, providing a clear method for assessing progress in systematic generalization research. 

%% file: sections/limitations.tex
\paragraph{Scope of our findings.}
In this work, we have considered an aspect of systematicity that is concerned with embedded sentences as defined by~\citet{hadley1994systematicity}. There is no guarantee that our findings generalize to scenarios that are different from ours, but that still falls in under some definition of systematic generalization.

\paragraph{Use of synthetic data.}
As is commonplace in compositionality research, our work relies on synthetic data that is generated by a CFG\@. This is necessary to quantify the entropy and to get a controlled experimental setup where the distribution of component parts in the embedded sentences can be measured precisely. This is generally not tractable for real-world corpora. Consequently, our method is constrained to cases where fine-grained distributional information is obtainable. 

%% file: sections/acknowledgements.tex
We acknowledge Sigma2---the National Infrastructure for High-Performance Computing and
Data Storage in Norway---for providing access to the LUMI super-computer, part of the EuroHPC Joint Undertaking, hosted by CSC (Finland) and the LUMI consortium. We also want to thank Lilja Øvrelid, Erik Velldal, and David Samuel at the Language Technology Group for discussions and feedback on the manuscript.